\documentclass[preprint,12pt]{elsarticle}

\usepackage{amssymb}
\usepackage{amsmath}
\usepackage{multirow}
\usepackage{graphicx}

\journal{Biomedical Signal Processing and Control}

\begin{document}

\begin{frontmatter}



\title{GCA-ResUNet: Medical Image Segmentation Using Grouped Coordinate Attention}


\author{Jun Ding} 
\author[1,*]{Shang Gao}

\affiliation{organization={Jiangsu University of Science and Technology},
            addressline={No. 666 Changhui Road, Zhenjiang, Jiangsu, P.R. China}, 
            city={zhenjiang},
            postcode={212003}, 
            state={jiangsu},
            country={China}}

\cortext[*]{Corresponding author: Gao Shang, Email: gao\_shang@just.edu.cn}


\begin{abstract}
Accurate segmentation of heterogeneous anatomical structures is pivotal for computer-aided diagnosis and subsequent clinical decision-making. Although U-Net based convolutional neural networks have achieved remarkable progress, their intrinsic locality and largely homogeneous attention formulations often limit the modeling of long-range contextual dependencies, especially in multi-organ scenarios and low-contrast regions. Transformer-based architectures mitigate this issue by leveraging global self-attention, but they usually require higher computational resources and larger training data, which may impede deployment in resource-constrained clinical environments.
In this paper, we propose GCA-ResUNet, an efficient medical image segmentation framework equipped with a lightweight and plug-and-play Grouped Coordinate Attention (GCA) module. The proposed GCA decouples channel-wise context modeling into multiple groups to explicitly account for semantic heterogeneity across channels, and integrates direction-aware coordinate encoding to capture structured spatial dependencies along horizontal and vertical axes. This design enhances global representation capability while preserving the efficiency advantages of CNN backbones. Extensive experiments on two widely used benchmarks, Synapse and ACDC, demonstrate that GCA-ResUNet achieves Dice scores of 86.11\% and 92.64\%, respectively, outperforming a range of representative CNN and Transformer-based methods, including Swin-UNet and TransUNet. In particular, GCA-ResUNet yields consistent improvements in delineating small anatomical structures with complex boundaries. These results indicate that the proposed approach provides a favorable trade-off between segmentation accuracy and computational efficiency, offering a practical and scalable solution for clinical deployment.
\end{abstract}

\begin{highlights}
\item We identify a fundamental limitation of existing unified attention mechanisms in CNN-based medical image segmentation, which tend to overlook channel-wise semantic heterogeneity under multi-organ and low-contrast conditions.
\item We propose a plug-and-play Grouped Coordinate Attention (GCA) module that decouples channel-wise context modeling and integrates direction-aware global pooling to enhance global contextual representation with minimal architectural modification.
\item We develop an effective integration strategy that embeds GCA into the bottleneck residual blocks of a  ResNet50 backbone, leading to improved delineation of small anatomical structures and blurred boundaries.
\item Extensive experiments on multiple public medical image segmentation benchmarks demonstrate consistent and robust performance improvements over strong CNN-based and Transformer-based baselines.
\end{highlights}


\end{frontmatter}



\section{Introduction}

Medical image segmentation is a cornerstone of computer-aided diagnosis and therapy, as it enables accurate delineation of organs and lesions for diagnosis, treatment planning, and longitudinal monitoring. Despite substantial progress, reliable segmentation in routine clinical practice remains challenging due to low contrast, heterogeneous tissue appearance, ambiguous boundaries, and large variations in anatomical scale. These difficulties are further amplified in multi-organ settings, where small or thin structures are easily overwhelmed by dominant regions in both feature learning and optimization. In addition, real-world deployment demands models that can incorporate global contextual cues while retaining stable and efficient architectural designs suitable for practical clinical environments.

Convolutional Neural Networks (CNNs)~\cite{LeCun1989,Krizhevsky2012,shen2023triplet}, represented by U-Net~\cite{ronneberger2015unet} and its variants~\cite{gu2019cenet,li2023eresunetpp}, have become the prevailing paradigm for medical image segmentation. The encoder--decoder structure with skip connections effectively fuses multi-scale features and preserves local spatial details, providing a strong inductive bias for dense prediction. Nevertheless, standard convolutions operate with limited receptive fields and rely on stacked layers to approximate long-range interactions, which restricts explicit global context modeling. Consequently, CNN-based models may exhibit boundary blurring, structural discontinuities, and reduced robustness in low-contrast regions, particularly when segmenting small or elongated anatomical structures whose appearance is easily dominated by surrounding tissues.

To mitigate the locality limitation of convolutions, Transformer-based architectures~\cite{vaswani2017attention,shen2023git,shen2023pbsl} have been introduced into medical image segmentation, either as replacements for convolutional components or as hybrid designs such as TransUNet~\cite{chen2024transunet} and Swin-UNet~\cite{cao2022swinunet}. By explicitly modeling long-range dependencies and cross-scale interactions, these approaches can improve segmentation performance on complex anatomy. However, self-attention typically incurs high computational and memory costs, and Transformer-based models often require more training data to generalize well. These factors can hinder their applicability in resource-constrained or data-scarce medical settings. Moreover, their gains may be less stable under domain shifts, which frequently occur across scanners, protocols, and institutions in clinical practice.

An alternative and increasingly practical direction is to enhance CNN-based architectures with attention mechanisms that inject global contextual modulation while preserving the favorable inductive bias and efficiency of convolutions. Representative examples include Squeeze-and-Excitation (SE)~\cite{hu2018senet}, Convolutional Block Attention Module ~\cite{woo2018cbam}, and Coordinate Attention~\cite{hou2021coordinateattention}. These methods recalibrate feature responses along channel and/or spatial dimensions to improve discriminability. However, most existing attention designs generate a unified attention pattern across the entire channel space or spatial domain. Such a unified formulation implicitly assumes homogeneous feature semantics, which is often violated in medical segmentation where feature channels may correspond to diverse anatomical cues. In multi-organ and low-contrast scenarios, unified attention can bias learning toward dominant structures, thereby weakening the representation of small-scale targets and compromising boundary delineation. In addition, global cues derived from coarse aggregation may be insufficient to preserve fine-grained structural details in ambiguous regions.

Motivated by these observations, we posit that effective global context modeling for medical image segmentation should explicitly account for channel-wise semantic heterogeneity, rather than enforcing a single shared attention pattern. To this end, we propose \textbf{GCA-ResUNet}, an efficient hybrid segmentation framework that integrates a plug-and-play \textbf{Grouped Coordinate Attention (GCA)} module into a ResNet50 backbone. The key idea is to decouple channel-wise context modeling into multiple groups, enabling diverse global representations to be learned across channel subspaces. Within each group, we adopt direction-aware coordinate encoding through horizontal and vertical global pooling to capture long-range structural dependencies while retaining spatial sensitivity. We further leverage both average pooling and max pooling to extract complementary statistics, and employ a shared $1\times 1$ convolution to transform grouped features and restore their original dimensionality.
The proposed GCA module is embedded into the bottleneck residual blocks of ResNet50~\cite{he2016deep}, specifically after the third convolution and normalization layer and before residual summation. This placement allows global contextual modulation to operate on semantically rich features while preserving the stability and optimization benefits of residual learning. Extensive experiments on public benchmarks demonstrate that GCA-ResUNet consistently improves segmentation accuracy and boundary delineation, with particularly notable gains on small anatomical structures and regions with blurred boundaries. We evaluate both region overlap and boundary quality using Dice and HD95 metrics, validating the effectiveness and robustness of the proposed approach.

The main contributions of this work are summarized as follows:
\begin{itemize}
    \item We analyze a key limitation of unified attention mechanisms in CNN-based medical image segmentation, namely their tendency to overlook channel-wise semantic heterogeneity in multi-organ and low-contrast scenarios.
    \item We propose a lightweight and plug-and-play Grouped Coordinate Attention (GCA) module that decouples channel-wise context modeling and integrates direction-aware global pooling to strengthen global contextual representation with minimal architectural modification.
    \item We present an effective integration strategy that embeds GCA into ResNet50 bottleneck residual blocks, improving the delineation of small anatomical structures and ambiguous boundaries while maintaining architectural stability.
    \item Extensive experiments on multiple public benchmarks demonstrate consistent improvements over strong CNN- and Transformer-based baselines in both overlap and boundary metrics.
\end{itemize}

\section{Related Work}

\subsection{CNN-based Medical Image Segmentation}

Convolutional Neural Networks (CNNs) have long served as the dominant paradigm for medical image segmentation due to their strong locality inductive bias, parameter sharing, and effectiveness in modeling local spatial patterns~\cite{LeCun1989,Krizhevsky2012}. Among these approaches, U-Net and its variants remain representative architectures. Through a symmetric encoder--decoder design with skip connections, U-Net effectively fuses multi-scale features by combining high-level semantic information with low-level spatial details, achieving robust performance across a wide range of medical image segmentation tasks~\cite{ronneberger2015unet}.
Building upon the U-Net framework, numerous studies have focused on enhancing representation capacity and optimization stability by adopting stronger backbone networks, such as ResNet-based or DenseNet-based encoders, and by redesigning feature fusion strategies~\cite{he2016deep,huang2017densely}. These improvements have led to more expressive and stable segmentation models. Nevertheless, CNN-based methods fundamentally rely on local convolution operations whose effective receptive fields are constrained by kernel size and network depth. As a result, long-range contextual dependencies are modeled only implicitly, which limits their ability to capture global anatomical relationships.
This limitation becomes particularly evident in medical image segmentation scenarios involving heterogeneous anatomical structures, low-contrast tissues, small or thin targets, and ambiguous boundaries. In such cases, CNN-based models often exhibit boundary blurring, structural discontinuities, or confusion between adjacent anatomical regions. These observations indicate that, while convolutional architectures remain highly effective, enhanced global context modeling is essential to address the intrinsic complexity of medical imaging data.

Recent work boosts segmentation by revisiting modern convolutional designs. For instance, ConvNeXt-inspired designs demonstrate that carefully optimized pure CNN architectures can achieve competitive performance with Transformer-based models by strengthening representation learning through improved convolutional blocks and normalization strategies~\cite{liu2022convnext}. In the medical imaging domain, MedNeXt further adapts modern ConvNet principles to encoder--decoder segmentation architectures and reports strong performance across multiple medical benchmarks~\cite{roy2023mednext}. In addition, progressive CNN-based segmentation frameworks have been proposed to iteratively refine predictions and improve robustness in complex anatomical scenarios, illustrating that convolutional pipelines can still be substantially enhanced through architectural refinement and contextual modeling~\cite{gong2023pimedseg}. Despite these advances, most CNN-based approaches continue to rely primarily on local convolutional interactions, leaving explicit global context modeling largely unaddressed.

\subsection{Transformer-Based Segmentation Networks}

To explicitly capture long-range dependencies beyond local convolutions, Transformer architectures~\cite{vaswani2017attention} have been increasingly adopted in computer vision and medical image segmentation. Representative U-shaped Transformer or hybrid architectures, such as TransUNet~\cite{chen2024transunet} and Swin-UNet~\cite{cao2022swinunet}, introduce self-attention to model global contextual interactions and cross-scale relationships, and have demonstrated strong performance on complex anatomical structures and multi-organ segmentation benchmarks.
Recent research has further investigated more structured attention patterns to mitigate the prohibitive cost of dense global self-attention. Window-based and hierarchical attention (e.g., shifted windows) reduce computation by restricting attention to local windows while maintaining cross-window information flow via stage-wise design~\cite{cao2022swinunet}. For volumetric segmentation, UNETR++ introduces an Efficient Paired Attention (EPA) block to jointly capture spatial and channel dependencies with improved practicality for 3D inputs~\cite{shaker2024unetrpp}. In addition, attention pattern redesigns such as cross-shaped windows aim to strengthen directional interactions (horizontal/vertical) that are highly relevant to anatomical structures, and have reported competitive results across multiple medical segmentation benchmarks~\cite{liu2025cswinunet}. Beyond classical Transformers, emerging state-space-model-based segmentation backbones (e.g., VM-UNet) have also been explored as alternatives for long-range modeling, reflecting the community’s interest in reducing quadratic attention bottlenecks while retaining global interaction capability~\cite{ruan2024vmunet}.

Despite these advances, Transformer-based segmentation networks still face notable practical challenges. First, self-attention is inherently expensive at high spatial resolution, and even window-based designs typically incur higher memory overhead than purely convolutional counterparts, especially for 3D volumetric inputs~\cite{vaswani2017attention,shaker2024unetrpp}. Second, due to weaker locality inductive bias, Transformers may require stronger regularization and larger datasets to train robustly; performance can degrade in data-scarce medical settings or under domain shifts commonly encountered in clinical practice~\cite{chen2024transunet,cao2022swinunet}. Third, tokenization and patch-based representations may compromise fine-grained boundary sensitivity if not carefully compensated by strong multi-scale feature fusion, which motivates hybrid designs and auxiliary context modules. Consequently, achieving effective global context modeling with strong robustness and practical resource requirements remains an active research direction.

\subsection{Hybrid Architectures, Attention Mechanisms, and Motivation}

A practical alternative to fully Transformer-based segmenters is to augment CNN backbones with lightweight attention modules, injecting global contextual cues while preserving the strong locality inductive bias and training stability of convolutions~\cite{oktay2018attentionunet,chen2024transunet,cao2022swinunet,gong2023pimedseg}. This hybrid design philosophy has been widely adopted in medical image segmentation, since it often offers a favorable accuracy--efficiency trade-off in resource-constrained clinical settings.
Representative plug-and-play attention mechanisms include Squeeze-and-Excitation (SE)~\cite{hu2018senet}, Convolutional Block Attention Module (CBAM)~\cite{woo2018cbam}, and Coordinate Attention (CoordAtt)~\cite{hou2021coordinateattention}. SE performs channel-wise recalibration via global pooling, CBAM extends this idea by coupling channel and spatial attention, and CoordAtt further injects direction-aware positional encoding through axis-wise pooling. These designs share a common modeling assumption: a {single unified attention pattern} is learned and applied to the full channel space (and/or spatial domain) within a layer. While effective in many settings, unified modulation can be suboptimal for medical images, whose feature channels often encode heterogeneous semantic factors across organs, boundaries, and fine structures. In multi-organ and low-contrast scenarios, unified attention may amplify dominant responses, weakening the representation of small or thin anatomical structures and ambiguous boundary regions, and thus leading to boundary blurring and inter-organ confusion.

This limitation resonates with broader observations in controllable generation, where explicitly disentangling heterogeneous factors and enabling specialized pathways is often crucial for fine-grained structure preservation and robust conditional modeling~\cite{shen2025imaggarment,shen2025imagdressing,shen2024imagpose,shen2024advancing,shenlong}. Motivated by these insights, we propose {Grouped Coordinate Attention (GCA)}, which relaxes the unified-attention assumption by partitioning channels into multiple groups and learning group-specific coordinate attention patterns. By allowing different channel subspaces to attend to complementary global context, GCA increases attention diversity, reduces cross-channel interference, and introduces a structural prior that encourages specialization. This grouped, coordinate-aware design aligns naturally with the heterogeneity of anatomical structures in medical images and provides a principled mechanism for enhancing global context modeling with minimal computational overhead.

\section{Methods}
\subsection{Main Network Structure}

\begin{figure*}[t]
    \centering
    \includegraphics[width=0.8\textwidth]{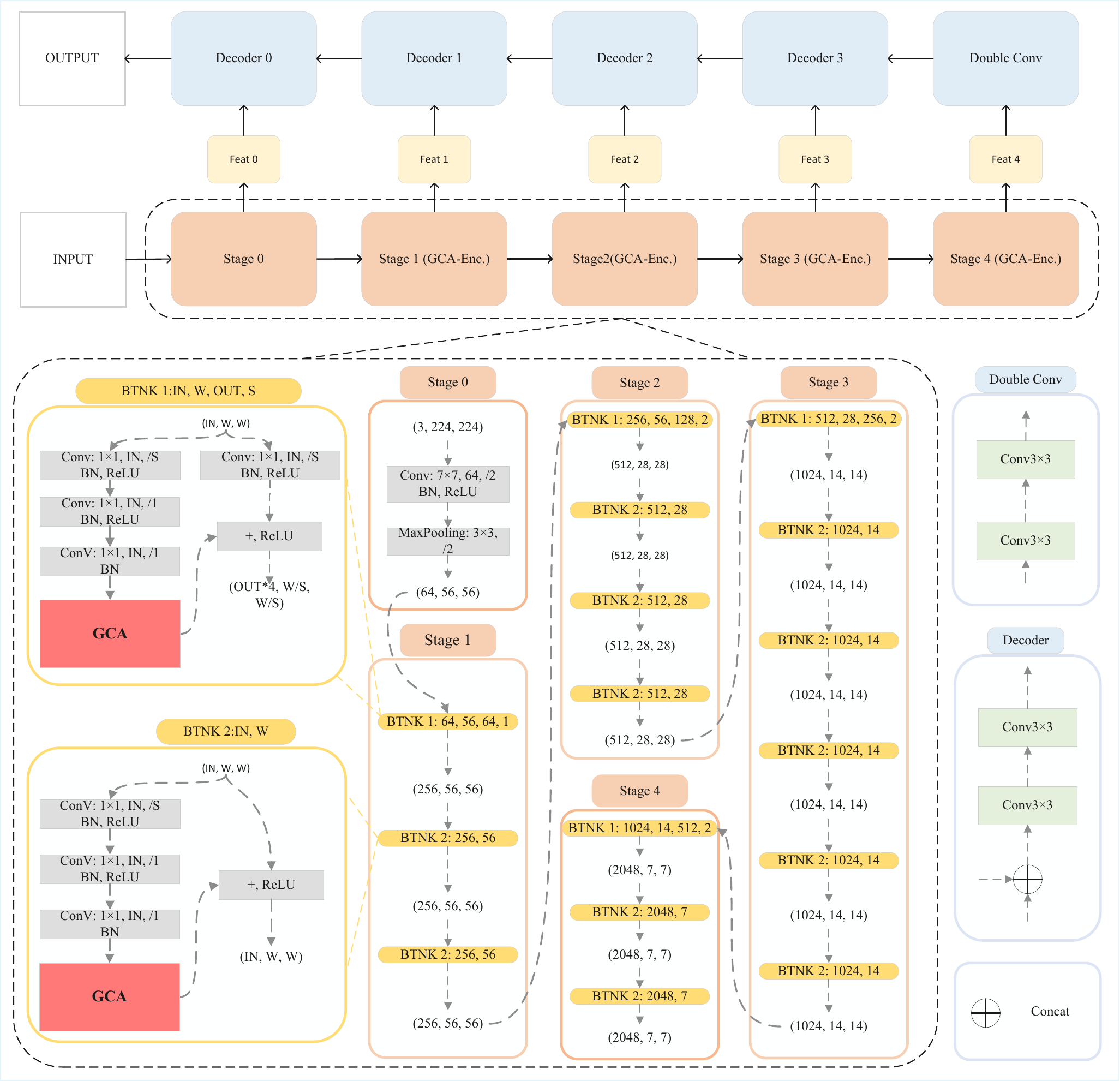}
    \caption{Schematic diagram of the GCA-ResUNet network. The architecture adopts a U-Net–style encoder–decoder structure, with residual blocks incorporated in the encoder and skip connections to enable multi-scale feature extraction and precise pixel-level segmentation.}
    \label{fig:GCA-ResUNet}
\end{figure*}

Fig.~\ref{fig:GCA-ResUNet} illustrates the overall architecture of the proposed GCA-ResUNet. The network follows the classical U-Net paradigm with a symmetric encoder--decoder structure and skip connections, which are effective for preserving spatial details during feature reconstruction. To enhance representation capacity while maintaining optimization stability, a ResNet50 backbone is adopted as the encoder, replacing the shallow stacked convolutional blocks in the original U-Net. The residual learning formulation enables deeper feature extraction and facilitates stable training, which is particularly beneficial for modeling complex anatomical patterns in medical images.

A key design choice of GCA-ResUNet is the integration of the proposed Grouped Coordinate Attention (GCA) module into the bottleneck residual blocks of the encoder. Specifically, GCA is inserted after the third convolution and batch normalization layer (\texttt{conv3 + BN}) and before residual summation within each ResNet50 bottleneck. This placement allows global contextual modulation to be applied to semantically rich feature representations, where channel-wise semantics are sufficiently abstracted, while preserving the residual learning pathway. As a result, long-range contextual dependencies and cross-channel interactions are enhanced without disrupting the locality inductive bias of convolutional operations.

The encoder begins with a $7\times7$ convolutional layer for initial feature extraction, followed by batch normalization and ReLU activation. A max-pooling operation is then applied to reduce spatial resolution. The network subsequently progresses through four residual stages, producing a hierarchy of multi-scale feature maps (Feat0--Feat4). Through progressive downsampling, deeper stages capture increasingly global contextual information, while shallower stages retain fine-grained spatial details. This hierarchical representation is essential for medical image segmentation tasks involving large anatomical scale variations and ambiguous boundaries.

The decoder adopts a U-Net-style upsampling strategy to gradually recover spatial resolution. At each decoding stage, low-resolution feature maps are first upsampled via bilinear interpolation and then concatenated with the corresponding encoder features through skip connections. The fused representations are refined using convolutional layers with ReLU activation, enabling effective integration of high-level semantic context and low-level spatial information. For the ResNet50 backbone, an additional convolutional layer is applied at the final decoding stage to further enhance feature aggregation prior to prediction.

Finally, a $1\times1$ convolutional layer projects the refined features to the target segmentation classes, enabling end-to-end pixel-wise prediction. Overall, the proposed architecture balances global contextual modeling and precise spatial localization. By combining residual learning, multi-scale feature fusion, and grouped coordinate attention, GCA-ResUNet is well suited for medical image segmentation scenarios characterized by heterogeneous anatomical structures and ambiguous boundaries.

\subsection{Backbone}
\label{sec:backbone}

Residual Networks (ResNets) are a foundational architecture in deep learning, largely due to identity-based skip connections that facilitate gradient propagation and stabilize the optimization of very deep models. By learning residual mappings, a bottleneck residual block can be formulated as
\begin{equation}
\mathbf{y} = \mathcal{F}(\mathbf{x}; \{W_i\}) + W_s \mathbf{x},
\end{equation}
where $\mathbf{x}$ and $\mathbf{y}$ denote the input and output feature maps, respectively; $\mathcal{F}(\cdot)$ represents the residual transformation implemented by a sequence of convolution and normalization layers; and $W_s$ is the shortcut mapping, which is an identity mapping when feature dimensions match and a projection (e.g., a $1\times1$ convolution) otherwise.

In this work, we adopt ResNet50 as the backbone encoder and adapt it for dense prediction. The original ResNet50 is designed for image classification and collapses spatial information into a single global representation via global average pooling, which is effective for large-scale classification benchmarks such as ImageNet~\cite{deng2009imagenet}. In contrast, dense prediction tasks (e.g., semantic segmentation, object detection, OCR, and medical image analysis) require spatially resolved, multi-scale representations. This requirement is particularly critical for medical image segmentation, where anatomical structures exhibit substantial scale variation and boundaries can be weak or ambiguous, making single-scale global features insufficient.

To address this, we remove the global pooling and classification head, and retain intermediate feature maps from the stem and each residual stage to form a hierarchical feature pyramid. As illustrated in Fig.~\ref{fig:GCA-ResUNet}, the encoder outputs five feature maps
\[
\{\text{Feat0},\text{Feat1},\text{Feat2},\text{Feat3},\text{Feat4}\},
\]
where $\text{Feat0}$ is extracted from the stem (initial convolution, normalization, and activation), and $\text{Feati}$ ($i=1,\ldots,4$) denotes the output of the $i$-th residual stage. These multi-level features provide rich semantic information at deep stages and fine spatial details at shallow stages, which are essential for accurate dense segmentation.

This hierarchical design naturally aligns with multi-scale segmentation architectures such as U-Net and Feature Pyramid Networks (FPN)~\cite{lin2017fpn}, enabling effective fusion of fine-grained spatial details from shallow layers and high-level semantic context from deeper layers.

The modified backbone retains the original ResNet50 stem, including the initial $7\times7$ convolution, max-pooling operation, and four residual stages (Layer1--Layer4) composed of bottleneck blocks. For dense prediction, the global average pooling layer (\texttt{avgpool}) and the fully connected classification layer (\texttt{fc}) are removed, ensuring that spatially resolved feature maps are preserved throughout the network.

Importantly, the residual learning formulation provides a stable and expressive foundation for incorporating additional contextual modeling modules. By maintaining identity shortcuts and bottleneck structures, the backbone preserves favorable gradient propagation properties while producing semantically rich feature representations at multiple scales. This design makes ResNet50 particularly suitable as a backbone for segmentation architectures that require both precise spatial localization and robust high-level semantics.

Although the proposed backbone remains architecturally compatible with ImageNet~\cite{deng2009imagenet} pre-trained weights, all experiments in this study are conducted without external pretraining, and all parameters are randomly initialized. This setting enables a fair evaluation of the proposed architecture without confounding effects from large-scale pretraining.

Overall, the modified ResNet50 backbone outputs a set of multi-level feature maps $\{\text{Feat0}, \ldots, \text{Feat4}\}$ with rich spatial and semantic information, providing a strong and stable foundation for subsequent context-enhancement modules and downstream dense prediction tasks.

\subsection{Grouped Coordinate Attention}

Building upon the multi-scale and semantically rich feature representations provided by the backbone network, we propose Grouped Coordinate Attention (GCA), an attention mechanism designed to relax the unified attention assumption and explicitly model channel-wise semantic heterogeneity. In contrast to existing attention modules that generate a single shared modulation pattern across all channels, GCA introduces group-wise coordinate attention to enable diverse global context representations to be learned in parallel. This design is particularly suitable for medical image segmentation, where heterogeneous anatomical structures and imbalanced target scales require differentiated contextual modeling.

Let the input feature tensor be $X \in \mathbb{R}^{B \times C \times H \times W}$, where $B$, $C$, $H$, and $W$ denote the batch size, number of channels, height, and width, respectively. We first partition the channel dimension into $G$ groups, each containing $C_g = C / G$ channels. By decoupling the channel space into multiple subspaces, GCA encourages different groups to focus on complementary semantic patterns and reduces cross-channel interference inherent in unified attention mechanisms. For the $g$-th group feature $X_g \in \mathbb{R}^{B \times C_g \times H \times W}$, GCA performs direction-aware global pooling to capture long-range dependencies while preserving positional information.

Specifically, horizontal pooling aggregates contextual information along the width dimension while preserving height:
\begin{equation}
f^{h}_{avg} = \text{AvgPool}_h(X_g), \quad
f^{h}_{max} = \text{MaxPool}_h(X_g),
\end{equation}
where $f^{h}_{avg}, f^{h}_{max} \in \mathbb{R}^{B \times C_g \times H \times 1}$. Average pooling captures smooth and holistic contextual responses, whereas max pooling emphasizes salient activations and boundary-related cues along the horizontal direction.

Similarly, vertical pooling aggregates information along the height dimension:
\begin{equation}
f^{w}_{avg} = \text{AvgPool}_w(X_g), \quad
f^{w}_{max} = \text{MaxPool}_w(X_g),
\end{equation}
where $f^{w}_{avg}, f^{w}_{max} \in \mathbb{R}^{B \times C_g \times 1 \times W}$. These operations encode complementary vertical contextual patterns and spatial dependencies.

The pooled features from both directions are fused and passed through a shared bottleneck transformation composed of two $1 \times 1$ convolutions with batch normalization and ReLU activation. The first convolution reduces the channel dimension to $C_g / r$, and the second restores it to $C_g$, where $r$ denotes the reduction ratio. Formally, the attention generation process can be expressed as
\begin{equation}
A = \sigma \big( \text{Conv}_{1\times1} ( \delta ( \text{BN} ( \text{Conv}_{1\times1}(F) ) ) ) \big),
\end{equation}
where $F$ denotes the fused pooled features, $\delta(\cdot)$ is the ReLU activation function, and $\sigma(\cdot)$ is the Sigmoid function. The resulting attention tensor is subsequently split into horizontal and vertical components,
\begin{equation}
A^{h} \in \mathbb{R}^{B \times C_g \times H \times 1}, \quad
A^{w} \in \mathbb{R}^{B \times C_g \times 1 \times W},
\end{equation}
which encode direction-aware contextual dependencies.

The refined group feature is obtained by applying the attention maps to the original feature via broadcasted element-wise multiplication:
\begin{equation}
Y_g = X_g \odot A^{h} \odot A^{w},
\end{equation}
where $\odot$ denotes element-wise multiplication. Finally, the outputs of all groups are concatenated along the channel dimension to form the final output
\begin{equation}
Y = \text{Concat}(Y_1, Y_2, \ldots, Y_G).
\end{equation}

From a conceptual perspective, GCA generalizes coordinate attention by replacing the single shared coordinate-aware modulation with multiple group-specific attention patterns. Compared with SE, GCA avoids collapsing spatial dimensions into a single channel descriptor and preserves directional spatial information. Compared with CBAM, GCA eliminates the need for additional spatial convolution branches by relying on efficient axis-wise pooling. Compared with Coordinate Attention, GCA introduces channel-wise grouping, enabling different channel subspaces to capture complementary global cues that cannot be simultaneously emphasized by a unified attention map. From an inductive-bias standpoint, this grouped design encourages specialization across channel groups, reduces cross-channel interference, and acts as an implicit regularizer during attention learning.

When integrated into ResNet50 bottleneck blocks, GCA provides direction-aware global modulation immediately after local convolutional aggregation and before residual fusion. This placement enables global contextual cues and local features to be jointly refined prior to identity addition, enhancing sensitivity to small anatomical structures, blurred boundaries, and complex textures while preserving the optimization stability of residual learning. In essence, the novelty of GCA does not lie in a simple combination of grouping and coordinate attention, but in reformulating global attention from a unified modeling assumption to a factorized, group-specific representation. This shift enables multiple heterogeneous global cues to be encoded in parallel within a single layer, which is particularly critical for multi-organ medical image segmentation.

\begin{figure*}[t]
    \centering
    \includegraphics[width=1\textwidth,height=0.9\textheight,keepaspectratio,trim=0 0 0 0,clip]{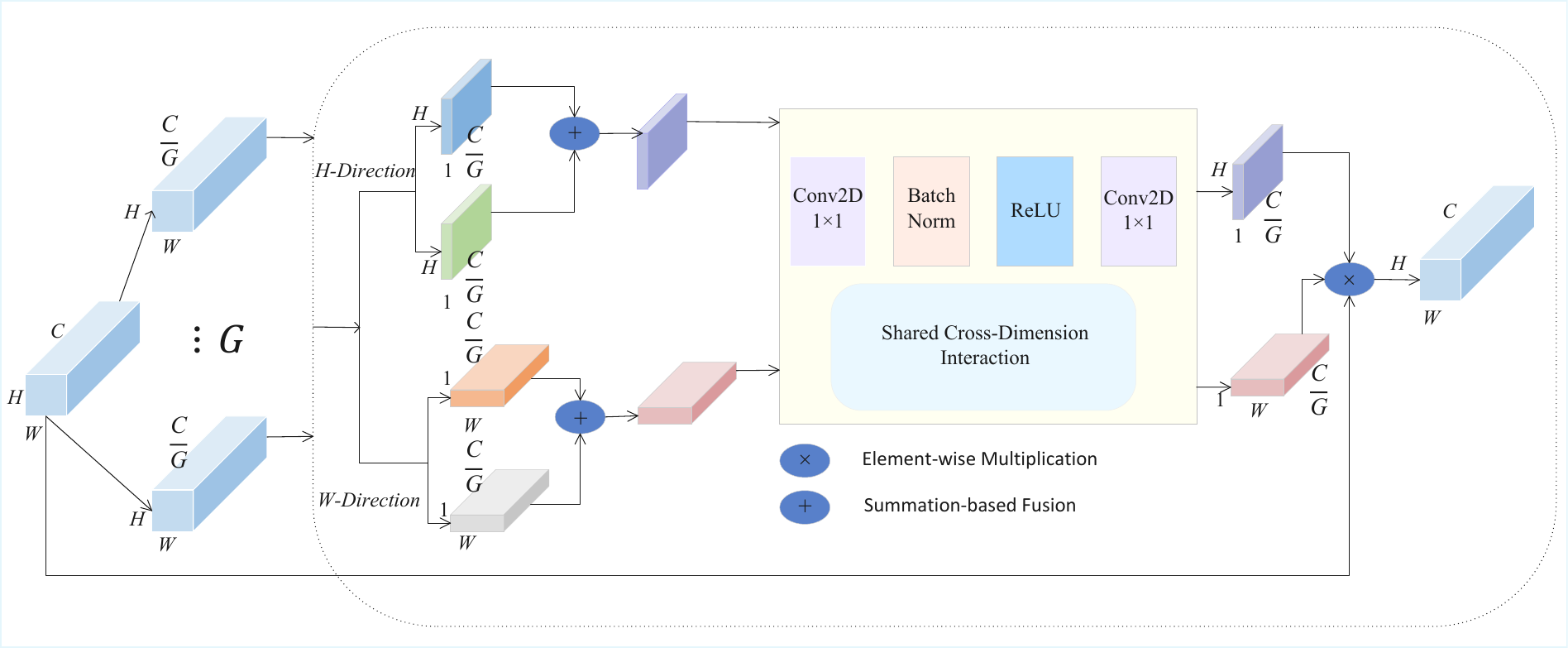}
    \caption{Grouped Coordinate Attention (GCA) network model diagram.}
    \label{fig:GCA}
\end{figure*}

\subsection{Decoder}

The decoder is designed to progressively recover the spatial resolution lost during encoder downsampling while effectively integrating multi-level feature representations. Following the U-Net paradigm, the decoder emphasizes precise spatial reconstruction by combining high-level semantic information with low-level structural details through skip connections.

At each decoding stage, the low-resolution feature map is first upsampled by a factor of two using bilinear interpolation to match the spatial resolution of the corresponding encoder feature. The upsampled feature is then concatenated with the encoder feature along the channel dimension, enabling the fusion of semantically rich representations from deeper layers and fine-grained spatial information from shallower layers.

The concatenated feature map is subsequently refined by two successive $3 \times 3$ convolutional layers with ReLU activation. The first convolution performs initial feature fusion and channel reorganization, while the second convolution further enhances local semantic consistency and preserves spatial continuity. This progressive refinement strategy improves the decoder’s ability to recover fine anatomical structures and delineate ambiguous boundaries, which are common challenges in medical image segmentation.

Bilinear interpolation is adopted to perform spatial upsampling, which provides smooth feature reconstruction and avoids spatial artifacts that may interfere with boundary recovery. By jointly leveraging multi-scale encoder features and progressive convolutional refinement, the decoder effectively reconstructs high-resolution segmentation maps that are consistent with anatomical structures in medical images.

\section{Experiment}
\subsection{Datasets}
\textbf{Synapse Multi-Organ Segmentation Dataset (Synapse).}
Synapse dataset is a widely used public benchmark for abdominal multi-organ segmentation. It contains 30 clinical abdominal CT scans with a total of 3,779 axial slices. All images are manually annotated at the pixel level by expert radiologists for eight abdominal organs: aorta, gallbladder, spleen, left kidney, right kidney, liver, pancreas, and stomach. Owing to low soft-tissue contrast, large anatomical variability, and ambiguous organ boundaries, Synapse poses significant challenges for segmentation models and is commonly used to evaluate robustness in multi-organ scenarios. We report the Dice Similarity Coefficient (DSC) as the primary metric and additionally provide the 95\% Hausdorff distance (HD95) for boundary accuracy. To reduce potential bias caused by the limited number of scans, we adopt a \textbf{four-fold cross-validation} strategy, and results on Synapse are reported as the mean $\pm$ standard deviation across the four folds.

\textbf{Automated Cardiac Diagnosis Challenge Dataset (ACDC).}
ACDC dataset consists of cine cardiac MRI scans from 100 patients, with expert annotations for three anatomical structures: left ventricle (LV), right ventricle (RV), and myocardium (Myo). We follow the \textbf{official ACDC split} for training, validation, and testing to ensure standardized evaluation and fair comparison with prior work. We report DSC as the primary metric and additionally provide HD95 to assess boundary accuracy. To account for randomness in training, we repeat experiments \textbf{three times} with different random seeds (10, 11, and 12) and report the mean $\pm$ standard deviation on the official test set.

\textbf{ISIC 2017 and ISIC 2018 Skin Lesion Segmentation Datasets.}
ISIC 2017 and ISIC 2018 datasets, released by the International Skin Imaging Collaboration (ISIC), are widely used benchmarks for skin lesion segmentation. Both datasets contain dermoscopic RGB images with pixel-level annotations provided by expert dermatologists. We follow the \textbf{official training/validation/test split} for both ISIC 2017 and ISIC 2018 and conduct experiments on the two datasets independently to evaluate generalization across different data distributions. All images are resized to \textbf{256$\times$256} during preprocessing. Following common practice in lesion segmentation, we report multiple metrics, including DSC and mean Intersection over Union (mIoU) for region overlap, as well as pixel-wise Accuracy (Acc), Specificity (Spe), and Sensitivity (Sen) to characterize false-positive and false-negative behaviors under class imbalance. For ISIC 2017/2018, experiments are repeated \textbf{three times} with different random seeds (10, 11, and 12), and the mean $\pm$ standard deviation on the official test sets is reported.

\begin{table}[t]
    \centering
    \caption{Performance comparison on the Synapse multi-organ dataset.}
    \label{tab:synapse_results}
    \resizebox{\textwidth}{!}{
    \begin{tabular}{l *{10}{c}}
    \hline
    \textbf{Methods} 
    & \textbf{Aorta} 
    & \textbf{Gallbladder} 
    & \textbf{Kidney(L)} 
    & \textbf{Kidney(R)} 
    & \textbf{Liver} 
    & \textbf{Pancreas} 
    & \textbf{Spleen} 
    & \textbf{Stomach} 
   & \textbf{mDSC (\%)$\uparrow$} 
   & \textbf{mHD95 (mm)$\downarrow$}
 \\
    \hline

    Res50 UNet~\cite{chen2024transunet}  
    & 87.74 & 63.66 & 80.60 & 78.19 & 93.74 & 56.90 & 85.87 & 74.16 & 77.61 & 37.68 \\

    Res50 Att-UNet~\cite{oktay2018attentionunet}  
    & 55.92 & 63.91 & 79.20 & 72.71 & 93.56 & 49.37 & 87.19 & 74.95 & 75.57 & 36.94 \\

    U-Net~\cite{ronneberger2015unet}   
    & 89.07 & 69.72 & 77.77 & 68.60 & 93.43 & 53.98 & 86.67 & 75.58 & 76.85 & 39.70 \\

    Att-UNet~\cite{oktay2018attentionunet}  
    & 89.55 & 68.88 & 77.98 & 71.11 & 93.57 & 58.04 & 87.30 & 75.75 & 77.77 & 36.20 \\

    MT-UNet~\cite{wang2022mtunet}   
    & 87.92 & 64.99 & 81.47 & 77.29 & 93.06 & 59.46 & 87.75 & 76.81 & 78.59 & 26.59 \\

    Swin-UNet~\cite{cao2022swinunet}     
    & 85.47 & 66.53 & 83.28 & 79.61 & 94.29 & 56.58 & 90.66 & 76.60 & 79.13 & 21.55 \\

    SelfReg-UNet~\cite{zhu2024selfregunet}   
    & 86.07 & 69.65 & 85.12 & 82.58 & 94.18 & 61.08 & 87.42 & 78.22 & 80.54 & 20.18 \\

    VM-UNet~\cite{ruan2024vmunet}            
    & 86.40 & 69.41 & 86.16 & 82.76 & 94.17 & 58.80 & 89.51 & 81.40 & 81.08 & 19.21 \\

    Mew-UNet~\cite{ruan2022mewunet}           
    & 86.68 & 65.32 & 82.87 & 80.02 & 93.63 & 58.36 & 90.19 & 74.26 & 78.92 & \textbf{16.44} \\

    Auto-Dyn. Conv. + Loc. Att.~\cite{wang2025autodynlocatt} 
    & 88.72 & 70.82 & 86.50 & 83.30 & 95.17 & 67.28 & 91.34 & \textbf{84.69} & 83.48 & 21.31 \\

    \hline
    GCA-ResUNet 
    & $\mathbf{89.95 \pm 0.42}$ 
    & $\mathbf{74.71 \pm 0.69}$ 
    & $\mathbf{93.38 \pm 0.38}$ 
    & $\mathbf{91.60 \pm 0.41}$ 
    & $\mathbf{95.54 \pm 0.34}$ 
    & $\mathbf{68.95 \pm 0.72}$ 
    & $\mathbf{92.60 \pm 0.36}$ 
    & $\mathbf{82.15 \pm 0.59}$ 
    & $\mathbf{86.11 \pm 0.49}$ 
    & $\mathbf{18.96 \pm 1.07}$ \\
    \hline
    \end{tabular}
    }
\end{table}

\begin{figure}[t]
    \centering
    \includegraphics[width=\linewidth]{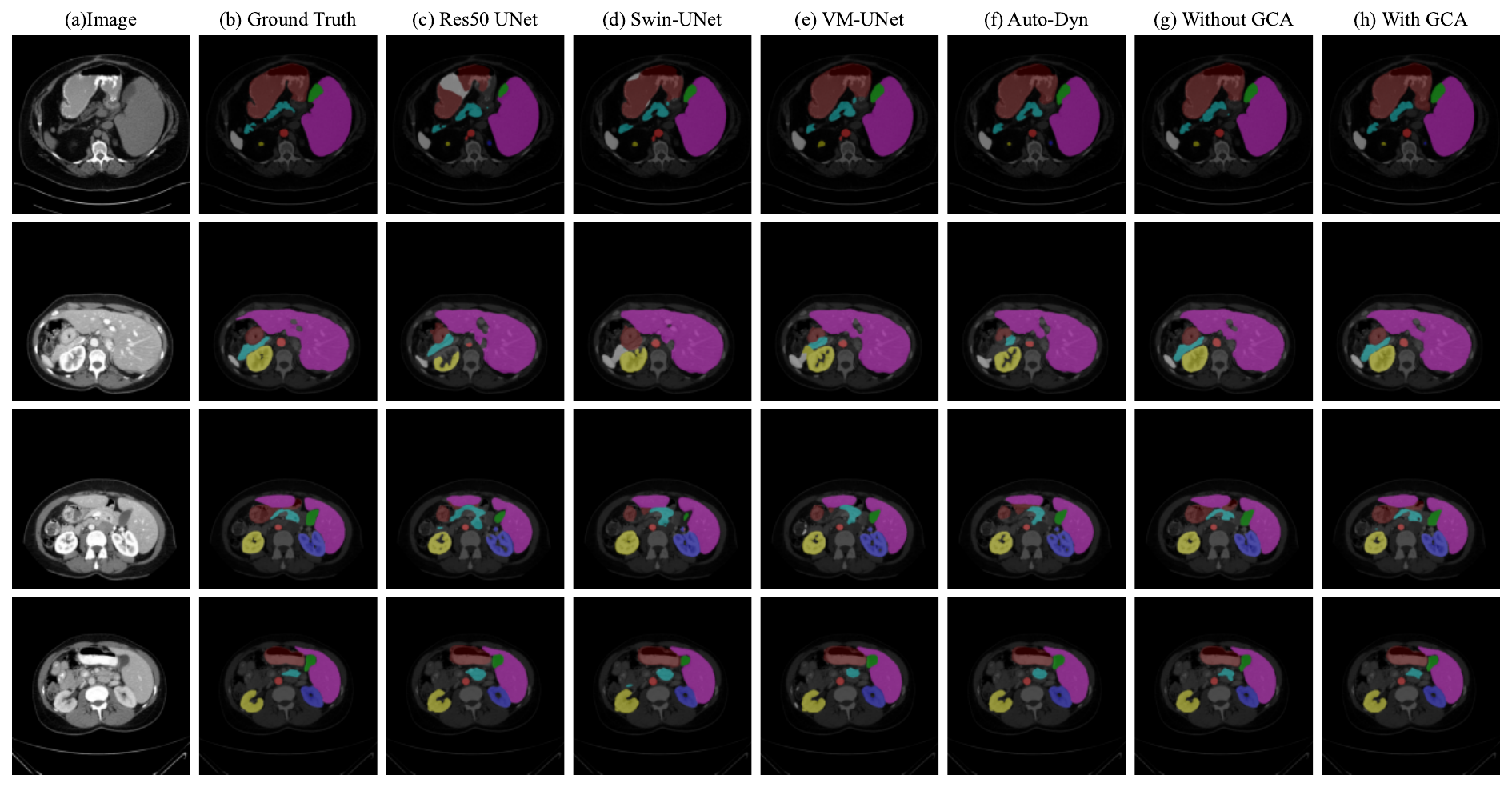} 
    \caption{Comparison of segmentation performance on the Synapse dataset.}
    \label{fig:output1}
\end{figure}
\begin{table}[!t]
    \centering
    \caption{Performance comparison on the ACDC dataset.}
    \label{tab:ACDC_results}

    \setlength{\tabcolsep}{1pt} 
    \small

    \begin{tabular}{l c c c c c}
        \hline
        \textbf{Methods} & \textbf{RV} & \textbf{Myo} & \textbf{LV} & \textbf{mDSC (\%)$\uparrow$} & \textbf{mHD95 (mm)$\downarrow$}
 \\
        \hline

        Res50 Att-UNet~\cite{oktay2018attentionunet} & 87.58 & 79.20 & 93.47 & 86.75 & 5.62 \\
        Res50 UNet~\cite{chen2024transunet}          & 87.10 & 80.63 & 94.92 & 87.55 & 2.44 \\
        U-Net~\cite{ronneberger2015unet}             & 87.17 & 87.21 & 94.68 & 89.68 & 2.61 \\
        Swin-UNet~\cite{cao2022swinunet}             & 88.55 & 85.62 & 95.83 & 90.00 & 1.60 \\
        MT-UNet~\cite{wang2022mtunet}                & 86.64 & 89.04 & 95.62 & 90.43 & 1.35 \\
        SelfReg-UNet~\cite{zhu2024selfregunet}       & 89.49 & 89.27 & 95.70 & 91.49 & 1.28 \\
        \hline

        GCA-ResUNet 
        & $\mathbf{92.27}$ 
        & $\mathbf{89.37}$ 
        & $\mathbf{96.30}$ 
        & $\mathbf{92.64 \pm 0.27}$ 
        & $\mathbf{1.09 \pm 0.13}$ \\
        \hline
    \end{tabular}
\end{table}

\begin{figure}[!t]
    \centering
    \includegraphics[width=\linewidth]{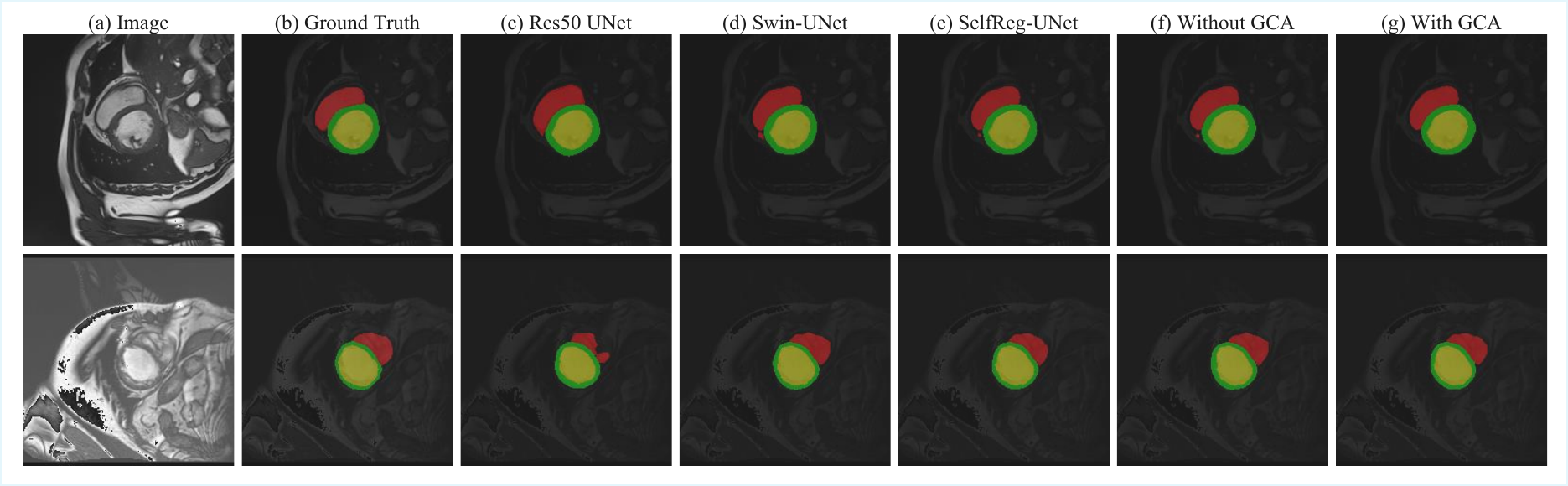} 
    \caption{Comparison of segmentation performance on the ACDC.}
    \label{fig:output2}
\end{figure}

\begin{table}[!t]
\centering
\caption{Performance comparison on the ISIC datasets.}
\label{tab:isic_results}
\resizebox{\textwidth}{!}{
\begin{tabular}{lcccccc}
\hline
\textbf{Dataset} & \textbf{Methods} & \textbf{mIoU(\%)$\uparrow$} &\textbf{DSC(\%)$\uparrow$} &
\textbf{Acc(\%)$\uparrow$} &\textbf{Spe(\%)$\uparrow$} &\textbf{Sen(\%)$\uparrow$}
 \\
\hline

\multirow{5}{*}{ISIC17}
& U-Net~\cite{ronneberger2015unet}            & 76.98 & 86.99 & 95.65 & 97.43 & 86.82 \\
& UTNetV2~\cite{gao2022utnetv2}              & 77.35 & 87.23 & 95.84 & 98.05 & 84.85 \\
& TransFuse~\cite{zhang2021transfuse}        & 79.21 & 88.40 & 96.17 & 97.98 & 87.14 \\
& MALUNet~\cite{ruan2022malunet}             & 78.78 & 88.13 & 96.18 & 98.47 & 84.78 \\
& GCA-ResUNet    & $\mathbf{79.91 \pm 0.34}$ & $\mathbf{88.93 \pm 0.28}$ & $\mathbf{96.81 \pm 0.21}$ & $\mathbf{98.58 \pm 0.41}$ & $\mathbf{89.90 \pm 0.32}$ \\
\hline

\multirow{8}{*}{ISIC18}
& U-Net~\cite{ronneberger2015unet}             & 77.86 & 87.55 & 94.05 & 96.69 & 85.86 \\
& UNet++~\cite{zhou2018unetplusplus}          & 78.31 & 87.83 & 94.02 & 95.75 & 88.65 \\
& Att-UNet~\cite{oktay2018attentionunet}      & 78.43 & 87.91 & 94.13 & 96.23 & 87.60 \\
& UTNetV2~\cite{gao2022utnetv2}               & 78.97 & 88.25 & 94.32 & 96.48 & 87.60 \\
& SANet~\cite{Wei2021SANet}                   & 79.52 & 88.59 & 94.39 & 95.97 & 89.46 \\
& TransFuse~\cite{zhang2021transfuse}         & 80.63 & 89.27 &94.66 & 95.74 & 91.28 \\
& MALUNet~\cite{ruan2022malunet}              & 80.25 & 89.04 & 94.62 & 96.19 & 89.74 \\
& GCA-ResUNet    & $\mathbf{81.05 \pm 0.29}$ & $\mathbf{89.41 \pm 0.21}$ & $\mathbf{94.67 \pm 0.25}$ & $\mathbf{96.75 \pm 0.38}$ & $\mathbf{91.39 \pm 0.29}$ \\

\hline
\end{tabular}
}
\end{table}

\subsection{Ablation Study}

To systematically evaluate the effectiveness of the proposed GCA module for medical image segmentation, we conduct comprehensive ablation studies based on a modified ResNet50 backbone. All experiments are performed under identical training settings to ensure fair comparisons. The Dice Similarity Coefficient (DSC) and the 95th percentile Hausdorff Distance (HD95) are adopted as the primary evaluation metrics, where higher DSC and lower HD95 indicate better segmentation performance. We first perform a module-level ablation study to analyze the impact of different architectural components and attention mechanisms. Here, modified backbone refers to the architecture described in Sec.~\ref{sec:backbone}, and w/o Attn denotes this modified backbone with all attention components removed. Specifically, we compare the original Res50 UNet with the modified backbone without any attention module (denoted as \textit{w/o Attn}) to examine the effect of backbone-level architectural refinements. Furthermore, SE (Squeeze-and-Excitation), CBAM (Convolutional Block Attention Module), and the proposed GCA are individually integrated into the same modified backbone to evaluate the contribution of different attention mechanisms under identical experimental settings.

\begin{figure}[t]
    \centering
    \includegraphics[width=\linewidth]{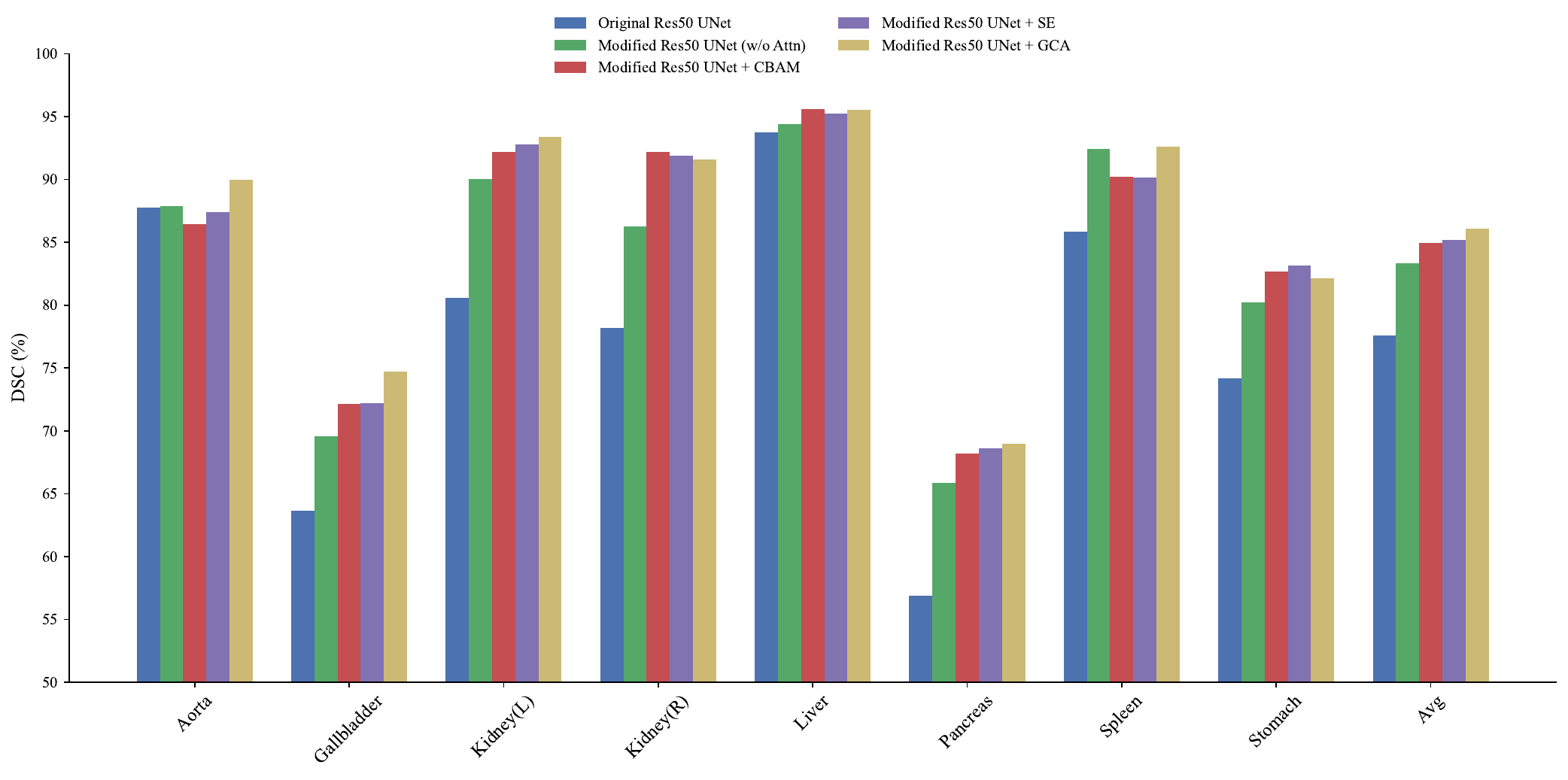}
    \caption{Performance comparison of the original and modified Res50 UNet equipped with different modules on the Synapse dataset in terms of DSC (\%). The y-axis is truncated to [50, 100] for better visualization.}
    \label{fig:Synapse_Ablation}
\end{figure}

\begin{figure}[t]
    \centering
    \includegraphics[width=\linewidth]{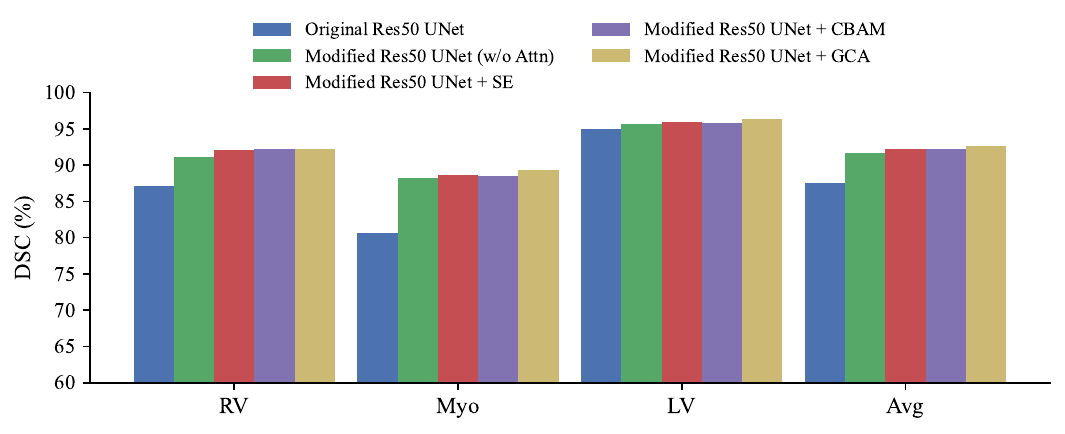}
    \caption{Performance comparison of the original and modified Res50 UNet with different modules on the ACDC dataset (DSC, \%). The y-axis is truncated to [60, 100] for better visualization.}
    \label{fig:ACDC_Ablation}
\end{figure}

The quantitative results on the Synapse and ACDC datasets are illustrated in Fig.~\ref{fig:Synapse_Ablation} and Fig.~\ref{fig:ACDC_Ablation}, respectively. As shown in these figures, the modified backbone without attention already achieves improved performance compared to the original Res50 UNet, validating the effectiveness of the architectural refinements. Moreover, incorporating attention mechanisms further enhances segmentation accuracy, among which the proposed GCA consistently outperforms SE and CBAM across different anatomical structures. This demonstrates that GCA is more effective in capturing global contextual information and modeling long-range dependencies, leading to more accurate and robust segmentation results.

In addition to segmentation performance, we quantify the computational overhead introduced by the proposed GCA module. Table~\ref{tab:complexity} reports model complexity measured with the Res50 UNet baseline (9 classes) under a 3-channel input of $224\times224$ and batch size $1$ on a single GPU. We report the number of parameters, MACs, FLOPs, and end-to-end inference latency under the same setting. Compared with the Res50 UNet baseline, integrating GCA increases the parameters from $43.93$\,M to $44.98$\,M, while the MACs increase from $17.57$\,G to $17.80$\,G (FLOPs from $35.15$\,G to $35.60$\,G). The measured latency changes from $6.83$ to $7.04$\,ms/img, indicating a small runtime overhead. Beyond the overall effectiveness of the GCA module, we further investigate the influence of its internal design choices through fine-grained hyperparameter ablation experiments. To avoid confounding factors introduced by dataset variability, all hyperparameter ablation studies are conducted on the Synapse dataset only, and both DSC and HD95 are reported for comprehensive evaluation.

\begin{table}[t]
\centering
\caption{Model complexity on Res50 UNet (9 classes) with $224\times224$ input and batch size 1 on a single GPU. FLOPs are approximated as $2\times$MACs.}
\label{tab:complexity}
\setlength{\tabcolsep}{4pt}
\small
\resizebox{\columnwidth}{!}{%
\begin{tabular}{lcccc}
\hline
\textbf{Model} & \textbf{Params (M)} & \textbf{MACs (G)} & \textbf{FLOPs (G)} & \textbf{Latency (ms)} \\
\hline
Res50 UNet (baseline) & 43.93 & 17.57 & 35.15 & 6.83 \\
Res50 UNet + GCA (Ours) & 44.98 & 17.80 & 35.60 & 7.04 \\
\hline
Overhead ($\Delta$) & +1.05 & +0.23 & +0.45 & +0.21 \\
\hline
\end{tabular}}
\end{table}

We first analyze the effect of channel grouping, which partitions feature channels into multiple groups to encourage diversified semantic representation and reduce cross-channel interference in global context modeling. In this experiment, the number of groups is varied while keeping other configurations unchanged. The results are summarized in Table~\ref{tab:ablation_groups}. As shown in Table~\ref{tab:ablation_groups}, introducing channel grouping leads to consistent improvements in both DSC and HD95 compared to the non-grouped variant. In particular, using two groups achieves the best balance between segmentation accuracy and boundary precision. Although increasing the number of groups further enforces stronger channel-wise specialization, excessive grouping slightly degrades performance due to insufficient interaction across semantic subspaces. Therefore, we set the number of groups to two as the default setting in our framework. We further investigate the effect of the reduction ratio, which controls channel compression in the attention branch and directly affects the representation capacity of the GCA module. In this experiment, we fix the number of groups to two and evaluate different reduction ratios on the Synapse dataset. The corresponding results are reported in Table~\ref{tab:ablation_reduction}. The results in Table~\ref{tab:ablation_reduction} indicate that a moderate reduction ratio yields the best overall performance in terms of both region overlap and boundary accuracy. While a smaller reduction ratio slightly increases model capacity, it introduces additional parameters with marginal improvements. Conversely, overly large reduction ratios excessively compress channel information, leading to inferior segmentation accuracy and degraded boundary delineation. Based on these observations, we fix the reduction ratio to 2 throughout all experiments, where $r$ denotes the reduction ratio. Finally, we analyze the contribution of different pooling strategies employed in the GCA module. Since both average pooling and max pooling are used to encode global contextual information, we compare three variants: average pooling only, max pooling only, and their combination. All experiments are conducted on the Synapse dataset, and the results are summarized in Table~\ref{tab:ablation_pooling}. As shown in Table~\ref{tab:ablation_pooling}, combining average pooling and max pooling achieves superior performance compared to using either strategy alone, resulting in higher DSC and lower HD95. Average pooling captures global contextual information, while max pooling emphasizes salient anatomical structures and sharp boundaries. Their combination provides complementary and more discriminative attention cues, which further improves segmentation accuracy and boundary delineation.

\begin{table}[t]
\centering
\caption{Ablation study on the number of channel groups in GCA on the Synapse dataset.}
\label{tab:ablation_groups}
\begin{tabular}{c c c c}
\hline
\textbf{Method} & \textbf{Group} & \textbf{DSC (\%)$\uparrow$} & \textbf{HD95 (mm)$\downarrow$} \\
\hline
GCA (no grouping) & 1 & 84.26 & 20.44 \\
GCA-G2 (Ours) & 2 & \textbf{86.11} & \textbf{18.96} \\
GCA-G4 & 4 & 85.81 & 21.07 \\
\hline
\end{tabular}
\end{table}

\begin{table}[t]
\centering
\caption{Ablation study on the reduction ratio in GCA on the Synapse dataset.}
\label{tab:ablation_reduction}
\begin{tabular}{c c c}
\hline
\textbf{Reduction Ratio} & \textbf{DSC (\%)$\uparrow$} & \textbf{HD95 (mm)$\downarrow$} \\
\hline
1 & 83.12 & 24.85 \\
2 (Ours) & \textbf{86.11} & \textbf{18.96} \\
4 & 86.05 & 20.44 \\
8 & 82.10 & 26.17 \\
\hline
\end{tabular}
\end{table}

\begin{table}[t]
\centering
\caption{Effect of different pooling strategies in GCA on the Synapse dataset.}
\label{tab:ablation_pooling}
\begin{tabular}{l c c}
\hline
\textbf{Pooling Strategy} & \textbf{DSC (\%)$\uparrow$} & \textbf{HD95 (mm)$\downarrow$} \\
\hline
Average Pooling only & 82.04 & 22.34 \\
Max Pooling only & 79.27 & 25.96 \\
Average + Max Pooling (Ours) & \textbf{86.11} & \textbf{18.96} \\
\hline
\end{tabular}
\end{table}

Based on the above ablation studies, we adopt GCA as the default configuration in our framework, using two groups, a reduction ratio of 2, and the combination of average and max pooling.

\subsection{Training}
All experiments were conducted under a unified and reproducible training protocol. For Synapse and ACDC, input images were resized to $224 \times 224$, while images from ISIC 2017 and ISIC 2018 were resized to $256 \times 256$. The batch size was set to 8. We trained all models using the AdamW optimizer with $\beta_1=0.9$ and $\beta_2=0.999$, weight decay of $1\times10^{-4}$, and an initial learning rate of $1\times10^{-4}$. A cosine learning-rate schedule was adopted to decay the learning rate from $1\times10^{-4}$ to $1\times10^{-6}$ over 200 epochs. Mixed-precision training was not used. For reproducibility, deterministic computation was enabled for all runs. For ACDC and ISIC 2017/2018, we repeated each experiment three times with different random seeds (10, 11, and 12) and report the mean $\pm$ standard deviation. For Synapse, we adopted four-fold cross-validation and report results aggregated across folds.

The training objective was defined as an unweighted sum of cross-entropy (CE) loss and Dice loss, i.e., $\mathcal{L}=\mathcal{L}_{\mathrm{CE}}+\mathcal{L}_{\mathrm{Dice}}$. We adopted modality-aware online augmentation. For Synapse and ACDC (CT/MRI), only geometric transformations were applied to preserve intensity-related semantics, including random scaling and mild aspect-ratio jittering (jitter factor 0.3; scale range $[0.8, 1.2]$), random horizontal flipping with probability 0.5, and random translation with padding to the target resolution (image padding value 128 and mask padding value 0). For ISIC 2017/2018 (RGB dermoscopy), the same geometric augmentations were used, and additional color perturbation was performed in the HSV space with hue/saturation/value jittering (hue 0.1, saturation 0.7, value 0.3). During validation and testing, images were resized using aspect-ratio preserving padding-based resizing. Input images were normalized by dividing pixel values by 255 before being fed into the network.

We applied an early-stopping strategy to prevent overfitting. The validation loss was monitored every 5 epochs, and training was terminated if the validation loss did not improve for 6 consecutive evaluations (patience = 6; i.e., 30 epochs). An improvement was defined as a decrease in validation loss of at least $1\times10^{-4}$ (min delta). The checkpoint achieving the lowest validation loss during training was retained for final testing.

All models were trained from scratch without using any pretrained weights, with parameters randomly initialized. Experiments were conducted on a single NVIDIA RTX 4060 Ti GPU (16 GB), with peak memory usage below 4 GB.

\section{Discussion}

This section provides a qualitative analysis of the observed experimental behaviors of the proposed method, with particular emphasis on its limitations and failure cases. Rather than reiterating quantitative performance gains, we focus on identifying scenarios in which the proposed approach may struggle and analyzing the underlying factors that contribute to such behavior.

\begin{figure}[t]
    \centering
    \includegraphics[width=\linewidth]{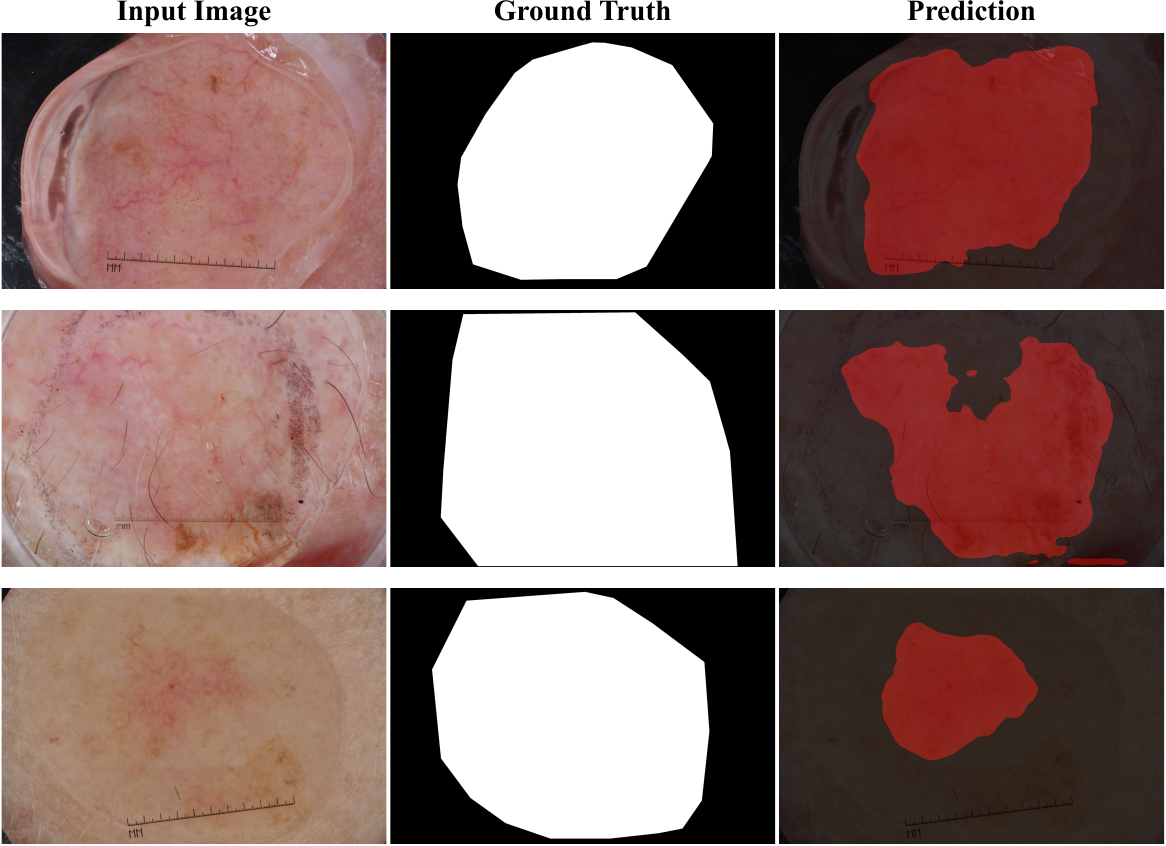}
    \caption{Representative failure cases of the proposed method in low-contrast regions. The model tends to produce inaccurate boundaries or partial omissions when foreground and background exhibit similar intensity and weak texture cues, highlighting the difficulty of segmenting anatomically shallow structures under ambiguous visual conditions.}
    \label{fig:failure}
\end{figure}

One notable limitation is observed in regions with low contrast or weak visual cues. As illustrated in Fig.~\ref{fig:failure}, anatomically shallow or lightly textured regions tend to exhibit inferior segmentation quality compared with areas characterized by clearer intensity or texture differences. This challenge becomes especially pronounced when the foreground and background share similar appearance characteristics. Although the proposed attention mechanism facilitates global contextual aggregation, reliably discriminating such ambiguous regions remains difficult when discriminative low-level features are insufficient. These observations suggest that global context modeling alone cannot fully compensate for the absence of strong local contrast in visually challenging regions.

Overall, the qualitative results indicate that the effectiveness of attention-based mechanisms is closely coupled with the quality of the underlying feature representations. While GCA enhances long-range dependency modeling and contextual awareness, its performance is ultimately constrained by the robustness of feature encoding in low-contrast scenarios. Improving segmentation accuracy under such conditions may therefore require complementary strategies, such as contrast-aware feature enhancement or more discriminative local representations.

In addition, all experiments in this study are conducted using models trained from scratch without external pretraining. Although this setting enables fair architectural comparison and isolates the contribution of the proposed design, it may limit the ability of the network to learn highly discriminative representations in challenging visual conditions. Leveraging large-scale pretraining or self-supervised representation learning could potentially improve robustness in low-contrast regions, and represents a promising direction for future work.

\section{Conclusion}

In this work, we present a systematic adaptation of the classical ResNet50 architecture for medical image segmentation. By refining the downsampling strategy and interface design, the backbone is seamlessly integrated with U-Net-style skip connections, and the proposed Grouped Coordinate Attention (GCA) module is embedded into bottleneck residual blocks to enhance long-range dependency modeling and global contextual awareness. This design enables the modified backbone to jointly capture fine-grained spatial details and high-level semantic information, resulting in improved boundary delineation and more accurate segmentation of small and challenging anatomical structures.
Extensive experiments conducted on multiple public benchmarks demonstrate that the proposed method consistently outperforms conventional ResNet50 and representative U-Net variants. The results validate the effectiveness of GCA in enhancing feature representation, contextual modeling, and robustness across different segmentation tasks and imaging modalities, including both CT and MRI data.
Overall, this study provides an effective enhancement of the ResNet50 backbone for semantic segmentation by explicitly addressing channel-wise semantic heterogeneity through grouped coordinate attention. Rather than relying on dense self-attention mechanisms, the proposed approach offers a structured and flexible way to enrich global context modeling within a convolutional framework. Future work will explore extending GCA to more advanced settings, including multi-task learning and three-dimensional medical image segmentation, as well as its integration with self-supervised or large-scale pretraining strategies to further improve generalization and practical applicability.

\bibliographystyle{elsarticle-num-names}
\bibliography{references}
\end{document}